\documentclass[sigconf, nonacm=true]{acmart}

\AtBeginDocument{%
  }

\usepackage{amsmath}
\usepackage{algorithm}
\usepackage{algorithmic}

\begin{document}

\title{Visual Product Graph: Bridging Visual Products And Composite Images For End-to-End Style Recommendations}

\author{Yue Li Du, Ben Alexander, Mikhail Antonenka, Rohan Mahadev, Hao-yu Wu, Dmitry Kislyuk}
\affiliation{%
  \institution{Pinterest, Inc}
  \city{San Francisco, CA}
  \country{USA}
}
  \email{{shirleydu, balexander, mantonenka, rmahadev, rexwu, dkislyuk}@pinterest.com}

\renewcommand{\shortauthors}{Pinterest Inc.}

\begin{abstract}
Retrieving semantically similar but visually distinct contents has been a critical capability in visual search systems. In this work, we aim to tackle this problem with Visual Product Graph (VPG), leveraging high-performance infrastructure for storage and state-of-the-art computer vision models for image understanding. VPG is built to be an online real-time retrieval system that enables navigation from individual products to composite scenes containing those products, along with complementary recommendations. Our system not only offers contextual insights by showcasing how products can be styled in a context, but also provides recommendations for complementary products drawn from these inspirations. 

We discuss the essential components for building the Visual Product Graph, along with the core computer vision model improvements across object detection, foundational visual embeddings, and other visual signals. Our system achieves a 78.8\% extremely similar@1 in end-to-end human relevance evaluations, and a 6\% module engagement rate. The "Ways to Style It" module, powered by the Visual Product Graph technology, is deployed in production at Pinterest.
\end{abstract}



\begin{teaserfigure}
  \includegraphics[width=\textwidth]{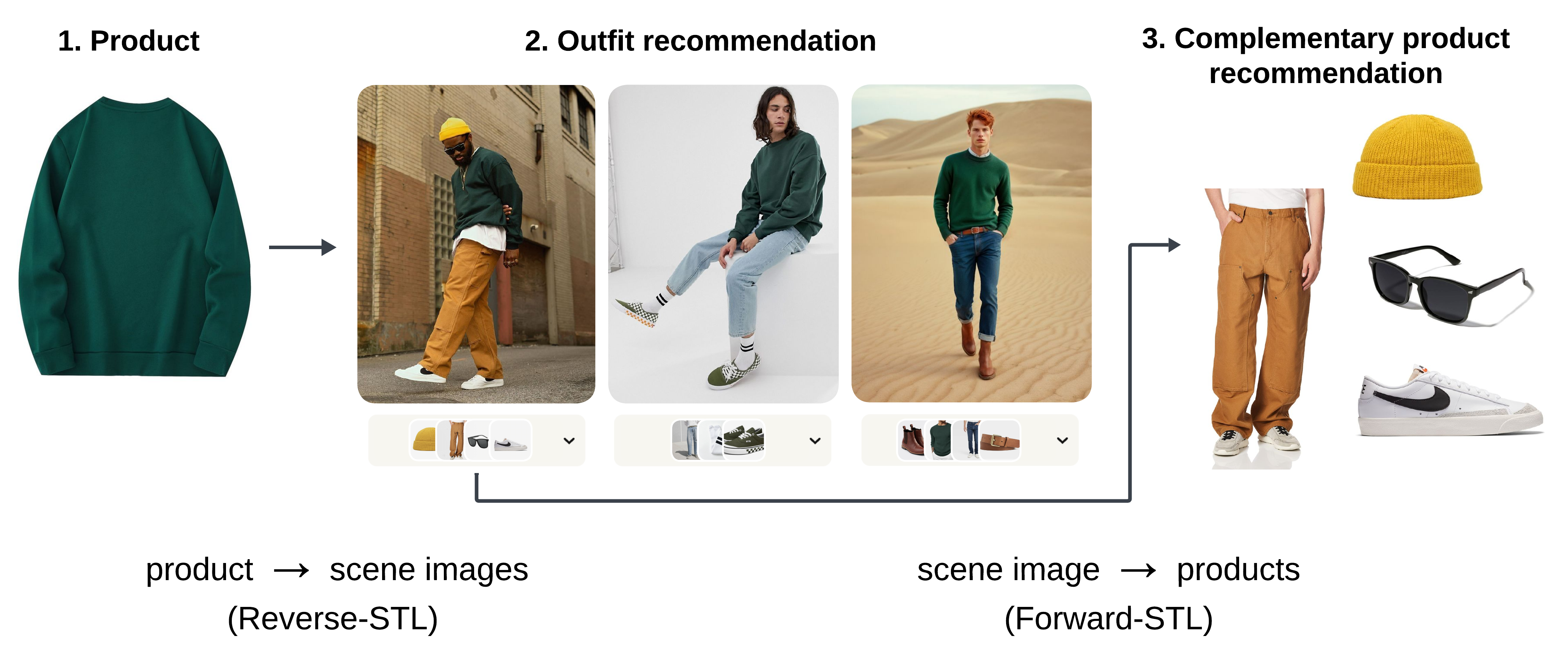}
  \caption{Visualization of the Visual Product Graph (VPG) system illustrating its dual capabilities (product-to-scene and scene-to-product) providing simultaneous outfit and complementary product recommendations from a single product input.}
  \Description{}
  \label{fig:teaser}
\end{teaserfigure}

\received{10 February 2025}

\maketitle

\begin{figure}
    \centering
    \includegraphics[width=1\linewidth]{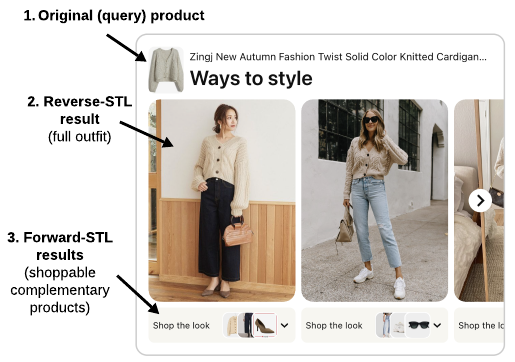}
    \caption{Pinterest's "Ways To Style It" module on Web. (1) The user is viewing the product page of the beige cardigan shown in the top left corner. (2) The module finds multiple full-outfit images containing this cardigan, to inspire the user about how to complete their outfit. (3) It displays shoppable versions of the other complementary items, allowing the user to purchase them}
    \label{fig:wtsi}
\end{figure}

\section{Introduction}
In modern e-commerce platforms, users often draw inspiration from both scene and product images. While existing solutions can recommend visually similar products when a user is inspired by a scene, they fall short when the inspiration comes from a standalone product image. In such cases, users typically seek guidance on how to style the product within a broader context. Furthermore, although many solutions exist for recommending complementary products, these are difficult to visualize within real-world settings and often lack social proof, leaving users uncertain about their purchasing decisions. As a result, there remains a significant gap in fully addressing these challenges within the e-commerce user experience.

To bridge the gap, we propose a novel solution: the Visual Product Graph (VPG). This framework provides a bidirectional mapping in styling contexts. It allows users to navigate from individual products to broader composite inspirational contexts, and vice versa. It showcases products in diverse styling scenarios and recommends complementary products based on user inspirations (Figure \ref{fig:teaser})). This functionality is applicable to both fashion and home decor domains. 

The Visual Product Graph leverages computer vision technologies, including object detection and visual embeddings. While existing literature has typically focused on scene-to-product lookup (i.e., traditional visual search), our approach uniquely introduces a reverse product-to-scene lookup, completing the loop between these two systems, termed \textbf{Forward-STL} (scene-to-product) and \textbf{Reverse-STL} (product-to-scene), respectively. This dual capability enhances user engagement by offering a more comprehensive and seamless shopping experience. Moreover, our method enhances complementary product recommendations beyond existing embedding-based methods by incorporating social proof (Section \ref{ctl_comparison}). 

This technology is deployed at Pinterest in the "Ways to Style It" module (Figure \ref{fig:wtsi}). In this module, users have the experience of exploring inspirational content that contains their product of interest. In this paper, we will dive into the design and implementation of how we built the system (Section  \ref{system_overview}). We will also show how we improved the computer vision models for improving the VPG -- specifically, object detection (Section \ref{object_detection}) and candidate retrieval (embedding) (Section \ref{embedding}) models -- through both modeling and dataset improvements. Our evaluation methodologies assess the impact of each component and demonstrate our ability to address specific VPG challenges. We showcase user-facing impacts through offline evaluations, human relevance assessments, and online A/B experiments (Section \ref{evaluation}) . We hope our contributions will give insights into how to connect the dots between products and inspirational images.

\section{Related Works}
\subsection{Shop the Look}
Visual search technology has been an important aspect of online shopping applications. The prior work "Shop The Look" \cite{stl} laid the groundwork for integrating visual search capabilities at Pinterest. When a user closes up on a scene image, an object detection model decomposes it into multiple objects, enabling the user to select an object of interest. Subsequently, an embedding model extracts the object visual embedding, which are then matched against our product corpus to retrieve visually similar, shoppable products. This object-to-products recommendation sets the stage for more advanced systems like the Visual Product Graph. Similar approaches in other contexts have also shown great results in semantic product search \cite{nigam2019semantic} \cite{li2021embedding} \cite{magnani2022semantic} and shopping recommendations \cite{baltescu2022itemsage}.
\subsection{Representation Learning}
Encoding real world entities like images, videos, audio, text, etc. into vectors allows us to perform meaningful tasks such as classification, recognition, and entity understanding. Learning these encodings to represent images, etc. is hence an essential task and is central to how modern deep learning based systems are able to condense information into lower-dimensional vectors \cite{lecun1998gradient}, \cite{hinton2006reducing}. Furthermore, these representations, or features, have shown great use in visual understanding problems including methods from classical computer vision using hand-crafted features \cite{lowe2004distinctive} as well as learned features based on CNNs \cite{krizhevsky2017imagenet}, and Vision Transformers \cite{dosovitskiy2020image}. Metric learning approaches \cite{hoffer2015deep} aim to learn these representations by distance comparisons. \cite{zhai2019learning} also describes how these representations can be used for solving visual search problems.
\subsection{Object Detection}
\begin{figure*}[h!]
    \centering
    \includegraphics[width=1\linewidth]{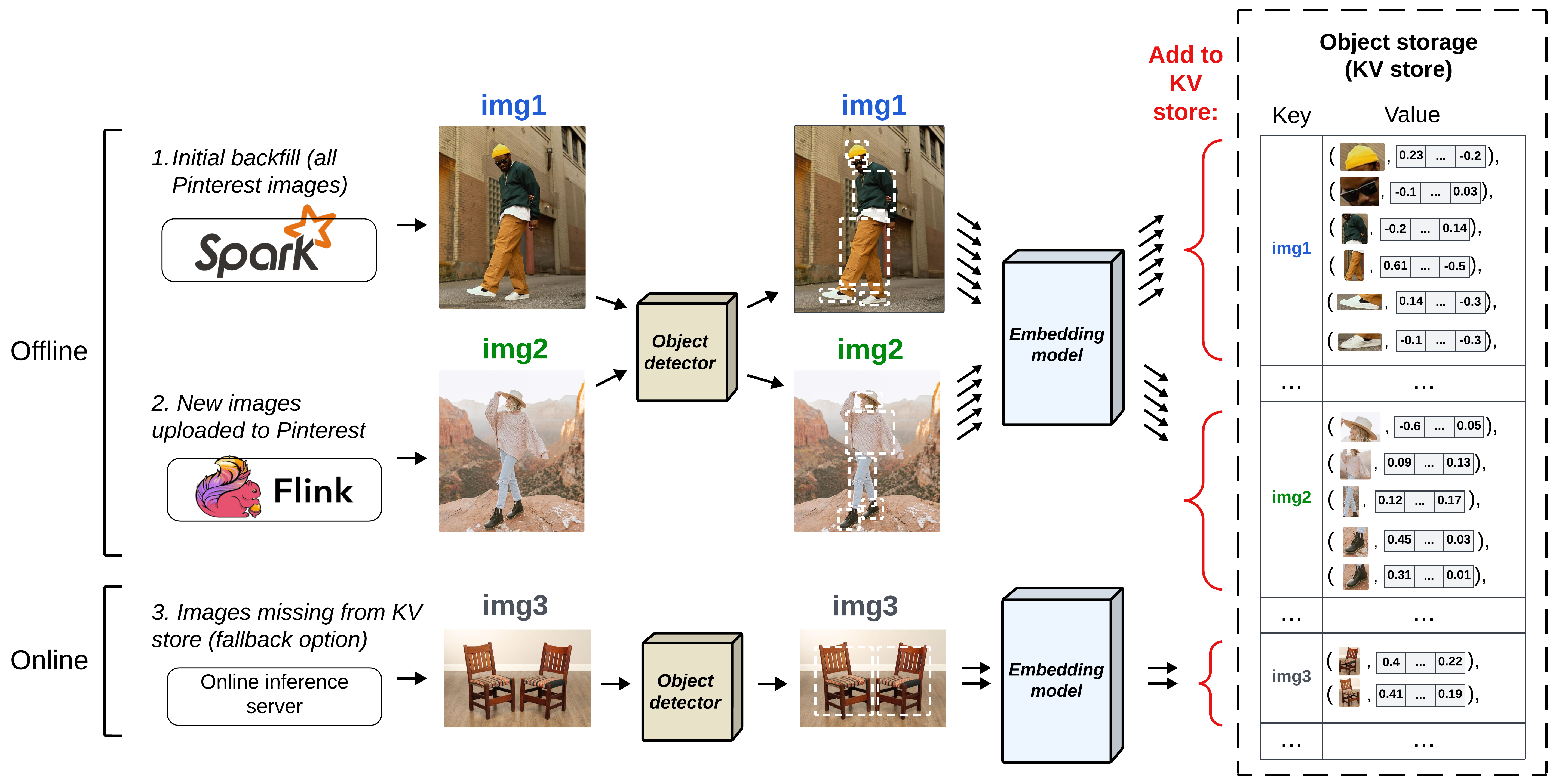}
    \caption{Leveraging online KV-Store to store objects and object embeddings. (1) A one-time backfill workflow is run offline in batches to backfill objects and their embeddings to the KV-store. (2) When new images are created on Pinterest, we use stream-processing service Flink to update the KV-store in near real-time. (3) As a fallback option, when users interact with an image whose key is not in the KV-store for some reason, we perform feature extraction online in real-time and write results to the KV-store. }
    \label{fig:gss}
\end{figure*}
Object recognition is crucial in visual shopping systems, enabling accurate product identification and categorization. Since the development of the YOLO (You Only Look Once) models \cite{yolo}, numerous enhancements have been introduced, with YOLOv8 marking the latest improvements in speed and accuracy. In recent years, transformer-based approaches have emerged as a strong alternative. DETR (Detection Transformer) \cite{detr} revolutionized object detection by directly predicting object sets using attention mechanisms. Further advancements, such as Co-DETR \cite{co-detr}, refine these methods by introducing collaborative training strategies to enhance detection performance.
\subsection{Complementary Products Recommendation}
In the domain of complementary product recommendations, numerous studies have sought to refine the methodologies for suggesting items that pair well with an initial product. Style modeling aims to better understand and predict visually cohesive sets of items. Techniques in this area have evolved from basic attribute matching to sophisticated machine learning models that analyze visual features and stylistic attributes to enhance recommendation accuracy, as explored by \cite{veit2015learningvisualclothingstyle}. Prior research \cite{ctl} represents a significant contribution to this field by introducing a triplet-based method. This approach harnesses the power of visual embeddings to assess stylistic compatibility between products, laying the groundwork for improved user engagement through more inspirational and contextually relevant recommendations. 
\section{System Overview}\label{system_overview}
In the Visual Product Graph system, there are two directions: from inspiration to products (Forward-STL), and from product to inspirations (Reverse-STL) (Figure \ref{fig:system_diagram}). In the Forward-STL path, we modify the existing Shop The Look (STL) \cite{stl} system slightly, which originally goes from object queries to products, so that it can now allow us to go from entire scene images to products. This path is described in Section \ref{stl-serving}. In the Reverse-STL path, we discuss how we provide inspirational recommendations given product queries. This is described in Section \ref{reverse-stl}. 
\subsection{Feature Storage} \label{object_storage}
In this section, we describe the infrastructure designed for the storage of full-image embeddings, object coordinates, and their corresponding embeddings. This billion-scale dictionary, encompassing full images, as well as decomposed objects and their embeddings, serves a dual purpose by providing significant benefits for both online serving and offline consumptions use cases.

We leverage Pinterest's large-scale online key-value store to construct this billion-scale key-value storage \cite{gss}. It operates through Rockstorewidecolumn, a wide-column, schemaless NoSQL database built on RocksDB. By leveraging RocksDB's efficient Log-Structured Merge (LSM) tree design and integrating it into a distributed system with sharding and replication, the database achieves high availability, scalability, and fault tolerance, supporting massive datasets and critical platform needs. 

To populate our objects signals in this storage, we set the keys to be image signatures, and the values to be lists of detected objects in the format of a dictionary: each object will have its XYWH coordinates, category, and embedding representation. There are two ways that we add values to this storage: (1) an initial backfill, and (2) incremental updates. This is illustrated in Figure \ref{fig:gss}. A similar process is done for populating the full-image signal. 
\begin{itemize}
    \item Initial backfill: we initiate the process with a one-time offline backfill workflow that performs object detection (Section~\ref{object_detection}) and then extracts unified visual embedding representations for each object (Section~\ref{embedding}), and subsequently stores these results within our storage. 
    \item  Incremental updates: the storage is updated in near real-time through a stream-processing framework based on Apache Flink, which facilitates the addition of new entries concurrent with image creation. Simultaneously, an online real-time system operates to update this storage with newly detected objects and embeddings as users interact with new images as a fallback in case the new image is not found in the storage yet. 
\end{itemize}

This key-value storage is instrumental not only in offline downstream usages (Section~\ref{object_index}) but also in reducing latencies during online serving (Section~\ref{rstl-serving}, \ref{stl-serving}). By enabling online access through key-value lookups, we effectively minimize feature computation costs. 
\subsection{Reverse-STL} \label{reverse-stl}
In this section, we describe the components of Reverse-STL, which facilitates the transition from individual product images to broader inspirational contexts. Specifically, Reverse-STL enables the retrieval of multiple composite images that feature the same or visually similar products from a given initial product image. These composite images possess higher inspirational value for Pinterest users compared to the original product images, which typically feature the item isolated against a plain white background. In Figure \ref{fig:wtsi}, our "Ways to Style It" screenshot shows a beige cardigan query and we offer styling ideas for similar cardigans in different styles and occasions via the "Reverse-STL" carousel. Additional inspirations are available by swiping through the carousel. Furthermore, the composite images offer "social proof" by demonstrating the item as part of an ensemble that others perceive as inspirational.

\begin{figure}[h]
    \centering
    \includegraphics[width=1\linewidth]{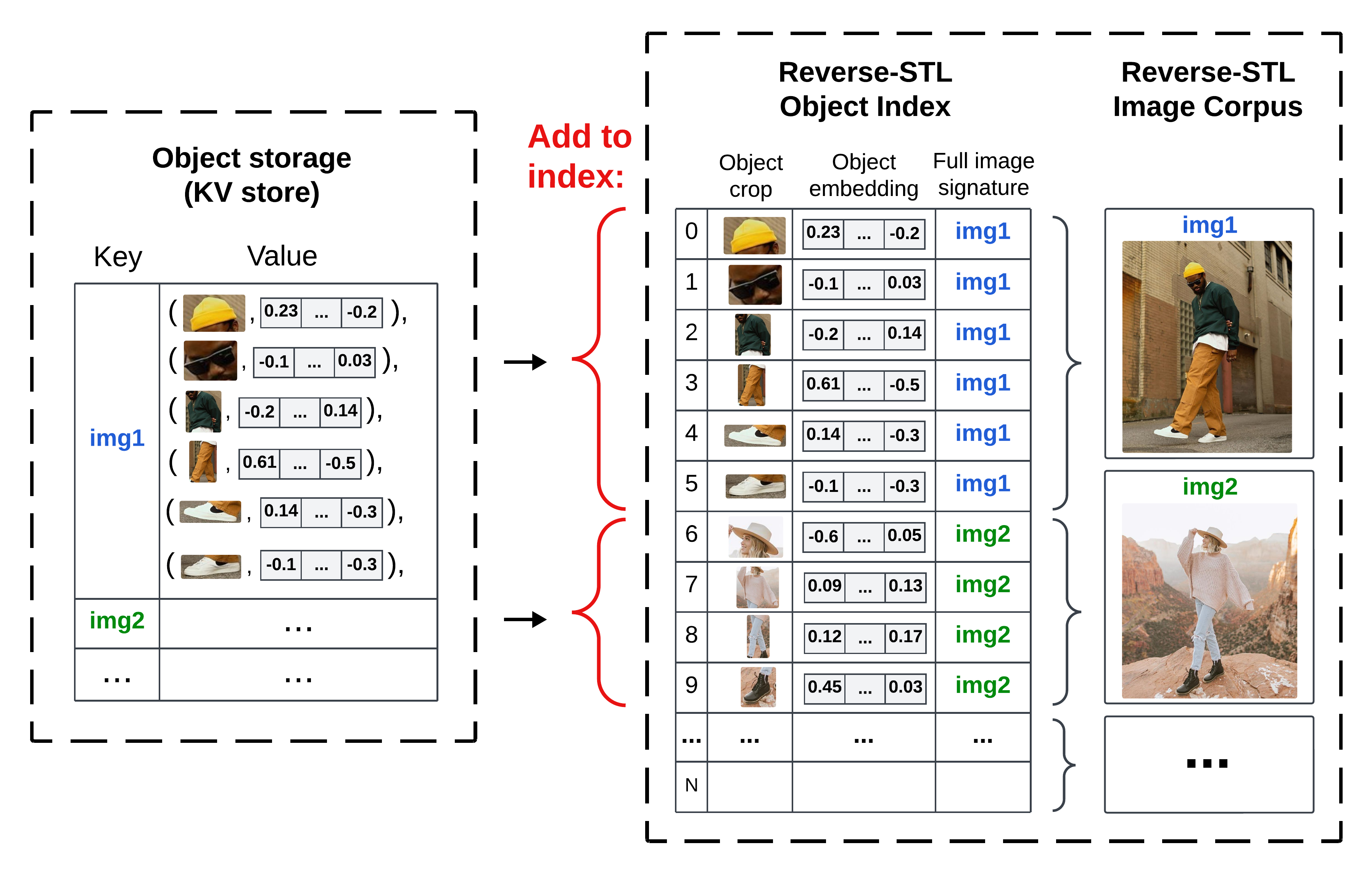}
    \caption{Building an object index for Reverse-STL offline. We extract the entries from the object storage and reformat them
to be indexed based on objects.}
    \label{fig:object_index}
\end{figure}

With the feature storage built from the previous step, we will now detail the methodology for constructing the object index in section \ref{object_index}, and we will detail the online serving challenges and methodologies in section \ref{rstl-serving}.

\subsubsection{$\boldsymbol{Object\ Index}$} \label{object_index}
The Reverse-STL index constitutes an offline object index that establishes a mapping from detected objects to their respective original composite images. We decompose the entries from the feature storage and reformat them to be indexed based on objects rather than images (Figure \ref{fig:object_index}). 

The object index consists of images that convey inspirational content by showcasing complete contexts within authentic backgrounds. These images must meet the criteria of having high resolution and visual appeal, ensuring their suitability for inclusion in the corpus. To ensure the corpus's high quality, we employ a series of filters based on proprietary signals derived from our unified embedding model (Section~\ref{embedding}). These filtering processes are visualized in \ref{object_index_filtering}. 
\begin{itemize}
    \item "Inspirational" filter:  Ensures each image contains a complete outfit or home decor scene.
    \item "Image quality" filters: Eliminates black-and-white images, collages, screenshots, and low-resolution or blurry images.
    \item "Object + shoppability" filters: Retains only images with objects that have high confidence scores, are sufficiently large, belong to allowed shopping categories, and contain objects from at least three shopping categories for broad shoppability.
\end{itemize}

Following this process, the corpus is refined to approximately 200 million high-quality, inspirational fashion images, encompassing roughly 900 million objects, and around 190 million home decor images, featuring approximately 1.8 billion objects. Finally, we maintain our index using Pinterest's Manas indexing system \cite{manas}.
\begin{figure*}[h!]
    \centering
    \includegraphics[width=1\linewidth]{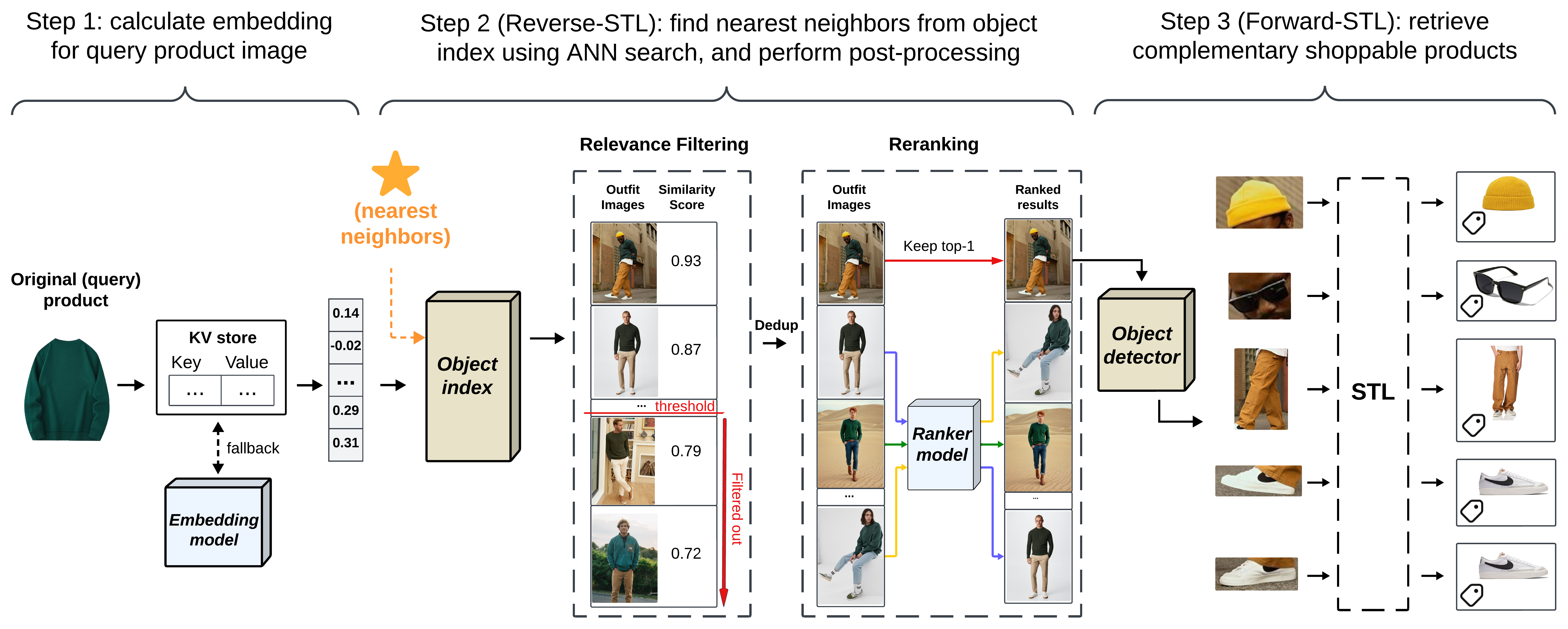}
    \caption{Overview of the Visual Product Graph system during serving.}
    \label{fig:system_diagram}
\end{figure*}
\subsubsection{$\boldsymbol{Serving}$} \label{rstl-serving}
We serve the Visual Product Graph in the "Ways to Style It" application.  Figure \ref{fig:system_diagram} provides an overview of the system during serving. First, the query product image's embedding is retrieved from the online key-value store. Next, we employ an HNSW-based \cite{hnsw} Approximate Nearest Neighbor (ANN) search engine \cite{manas} to facilitate the embedding-based retrieval process. This procedure aims to identify scene images containing objects that exhibit visual similarity to the query product. On average, the retrieval process returns approximately 150 candidate images. Subsequently, we apply a number of post-processing steps to ensure quality, as illustrated in Figure \ref{fig:system_diagram}. 
\begin{itemize}
    \item A relevance filter discards scene images with similarity scores below a set threshold, determined by the 75th percentile of the "top-5" candidate score distribution, thus eliminating less relevant results.
    \item To enhance diversity, we perform a de-duplication process using exact and near-duplicate \cite{neardup} image signatures. After all the post-processing steps, we are left with an average of 9.9 unique scene images.
    \item In the final phase, a re-ranking model reorders the images to optimize engagements. The top candidate remains unchanged to ensure high relevance, while the others are re-ranked for improved diversity and engagements.

\end{itemize}
\subsection{Forward-STL}\label{stl-serving}
In the Forward-STL path, we provide complementary product recommendations to the users. As illustrated in Figure \ref{fig:wtsi}, we provide complementary products for each of the inspirational results. By having this Forward-STL path, we enable users to complete the journey of being inspired by an outfit given a product they have in mind, and to complete the full outfit by being able to purchase the rest of the items in the outfit. The same applies to home decor setups. 

We build Forward-STL on top of the Shop The Look (STL) system \cite{stl}. In STL, the query is an object and we return visually similar product images to the query object. In our Forward-STL, the query is an image containing multiple objects, and we return visually similar product images to each of the decomposed objects in a round-robin fashion. We iterate through all the objects in each inspirational image to get the top 1 result from each of the objects. 

For building out the retrieval index, we maintain a corpus of products that are trustworthy. To be considered trustworthy, a product must be in-stock, from a legitimate website, and be a safe product. This trustworthiness check ensures that users will have a safe and reliable experience at Pinterest when actually deciding to make a purchase. The corpus size is 3B images. We have a daily workflow that adds incremental product images to the index.

During serving, the process selects only the top five inspirational results from Reverse-STL for Forward-STL retrieval due to its resource-intensive and latency-sensitive limitations. To further reduce latency, a cacheable client is employed for STL result retrieval, utilizing asynchronous and parallel fetching to address only non-cached scene pins. The cache entries are set with a two-hour Time-to-Live (TTL) to keep product information fresh and relevant to current offerings, thus minimizing stale data issues. The caching mechanism employs a key-value structure, where the key is based on the inspirational image signature and user details such as gender and country, while the values store the STL results for each image.

When a query, which is an inspirational result from the previous Reverse-STL step, comes in, detected objects and their embeddings are fetched from the online key-value store (Figure \ref{fig:gss}). Each inspirational image undergoes scene decomposition to identify up to four distinct objects within each image, selected based on the highest object confidence scores. The average hit rate to the key-value store is 99.92\%, enabling a significant reduction in the P99 latency of signal retrieval by 20.5 times and an eightfold reduction in model serving costs. This allows us to give users a seamless experience to load the complementary products without having to wait for the module to load.

The embedding-based retrieval for detected objects in each image is performed in batch mode, and the matching product images for each object are identified using the STL method \cite{stl}, yielding up to 12 potential candidates per image. Unsafe candidates and those belonging to non-matching categories (such as non-fashion items) are filtered out then. The remaining results are organized in a round-robin fashion, with only the top three results being retained and displayed for each inspirational image.

\section{Evaluation Methodology}\label{evaluation}
We evaluate our Visual Product Graph system using human relevance assessments for both the product-to-inspiration and inspiration-to-complementary products pathways. We also assess the "Ways to Style It" module's user engagements through online A/B testing. We also conduct offline evaluations for each underlying ML models to ensure their qualities. 
\begin{table}
    \centering
    \scalebox{0.69}{
     \begin{tabular}{c|c|l|l|l} \hline
 &Extremely Similar @ 1&Extremely similar @ 5&Similar @ 1 &Similar @ 5\\ \hline 
         Fashion&   31.8\%&26.2\%&78.8\%&76.3\%\\ \hline
 Home Decor& 18.7\%&17.3\%&61.1\%&60.3\%\\\end{tabular}
    }
        \caption{Visual Shopping end-to-end retrieval performance using the latest object detection model.}
    \label{tab:rstela_human_eval}
   
\end{table}

\textbf{End-to-End Extremely Similar (ES) Evaluation}: This involves measuring Precision@1 and Precision@5, indicating the proportion of retrieved results in the top-K that are extremely similar to the query. The ratings criteria are the same as the ones used in STL \cite{stl}:
    (1) Extremely similar: almost exactly the same.
    (2) Similar: not exactly the same, but a very good substitute in this context.
    (3) Marginally similar: don't look similar, but the same type of clothing and usually only share 1-3 key commalities.
    (4) Not similar: bad result. Mostly wrong type of clothing or the box includes wrong part of clothing.
    (5) Did not Load -- at least one Pin doesn't load or can't tell what a Pin contains.
For the product-to-inspiration pathway, we compile a dataset of the 5,000 most popular products on Pinterest based on user engagement. For each product, the system retrieves a scene image that may include the product object, subsequently verified by human raters to confirm the presence of the query product in the scene image. In this path, we evaluate fashion and home decor separately and the results are shown in Table \ref{tab:rstela_human_eval}. These relevance numbers are comparable to our STL path, which has been deployed in production for years. In evaluating the inspiration-to-complementary products pathway, we adopt the evaluation framework established in the Shop The Look \cite{stl} paper, thereby maintaining consistency in assessment methodologies.

\textbf{Module engagement rate}. We evaluate the module engagement rates with the Ways to Style It module. The module tap rate is calculated as \textit{module\ close up\ volume / module\ impression\ volume}, providing a quantitative measure of user interaction with the module. We observe a 6\% module engagement rate in this module. 

\textbf{Offline evaluations}. For our object detection model, we perform offline evaluations using mAP metrics to ensure detection accuracy. We also use R@P90 metrics to assess the balance between precision and recall in identifying relevant instances, optimizing for both high precision and comprehensive retrieval. Our embedding models are evaluated on retrieval accuracy, validating against known product matches to refine extremely similar quality. We will detail in later sections our designs of these offline evaluations.

\section{{Object Detection}}\label{object_detection}
The object detector plays a critical role in determining the coverage and quality of results within the Visual Product Graph. Consequently, we have focused on enhancing both accuracy and speed enhancements in our core object detection technology since the previous release of the Shop The Look \cite{stl} study:

(1) Improved detection performance via a target-collected training dataset.

(2) Upgraded from Faster R-CNN to YOLOv8 for enhanced speed and accuracy.

These enhancements in object detection have enabled the creation of a more refined and higher-quality object index. Consequently, the improvements significantly reduces latencies in the Forward-STL pathway, thereby enhancing the overall system efficiency and user experience. The overall improvements are summarized in Table \ref{tab:dtn_results}.
\subsection{Detection improvements}
\subsubsection{Target-collected training dataset}
We observed that collecting a dataset that's targeted for the most pressing issues helped lift the detection performance by a significant amount. Our original dataset comprises approximately 335k images, and we collected 60k images for this target-collected training dataset. It ensures diversity across various categories in fashion and home decor domains as well as demographic settings. There are 2 types of cases that are most pressing: 

(1) Important cases: We address failures among high-frequency STL queries. We perform human evaluations to verify the correctness of predicted categories. For queries where detection fails, these images serve as seeds to find visually similar examples using visual search methodologies.

(2) Difficult cases: We focus on challenging categories with low Recall@Precision metrics. By analyzing these alongside popular queries, we accumulate more examples to boost detection accuracy in these complex scenarios.

After collecting a targeted dataset, we address class imbalance issue by oversampling rare classes by a factor of $\sqrt{f_c/t}$, where $f_c$ is the frequency of the category, and $t$ is chosen as the 75th percentile of the data distribution. This approach ensures a balanced representation of the dataset across different categories.
\begin{table}
    \centering
    \begin{tabular}{c|c|l|l} \hline
 Model&mAP  &R@P90&Inference Speed \\ \hline 
         Production&   0.40 &0.24&81.89 ms/img\\ \hline
 + Target Dataset& 0.52 &0.33&81.89 ms/img\\ \hline 
 +Target Dataset + Co-DETR& $\boldsymbol{0.726}$ &0.423&172.53 ms/img\\ \hline 
         + Target Dataset + YOLOv8& 
     $\boldsymbol{0.69}$&$\boldsymbol{0.424}$&$\boldsymbol{33.48}$ ms/img\\ \hline\end{tabular}
    \caption{Visual Shopping end-to-end retrieval performance using the latest object detection model.}
    \label{tab:dtn_results}
\end{table}

\subsubsection{Model Architecture}
To address some of the common failure modes that we observed in our object detection pipeline, we upgraded our detector from a ResNext101-FasterRCNN \cite{fasterRCNN} with Feature Pyramid Network (FPN) \cite{fpn} model to a YOLOv8 model. We observed a 2.4x inference speed increase (81.89 ms/image -> 33.48 ms/image) with +32\% increase in mAP. We also observed that the YOLO model is better at detecting smaller objects and differentiating between duplicate objects, which is particularly helpful in home decor. By applying class-agnostic NMS, we reduced the amount of duplicate bounding boxes by about 50\% (Figure \ref{fig:dup_box} in Appendix).  It also helped reduce the false positive detections by almost 50\% while maintaining the same amount of precision (Figure \ref{fig:fp} in Appendix) . Additionally, we've explored another popular detection architecture, Co-DETR \cite{co-detr}. Although Co-DETR yielded slightly more mAP improvements, its inference speed was twice that of the production Faster R-CNN model, and is even 5x slower than our candidate YOLOv8 model, posing an expensive trade-off in computational efficiency. 
\subsection{Detection experiments}
We explore the following detection model variants (Table \ref{tab:dtn_results}):

\textbf{Baseline}. Faster-RCNN\cite{fasterRCNN} with a ResNext 32x8d FPN backbone

\textbf{+Target collected dataset}. Added the target collected dataset into our training dataset.

\textbf{+Co-DETR}. Upgraded the model to state-of-the-art Co-DETR model.

\textbf{+YOLOv8}. Upgraded the model to YOLOv8 model.

We see that our target-collected training dataset has brought +30\% in mAP and +37.5\% in R@P90 metrics. The modeling upgrade choosing the YOLOv8 model brough us another +33\% in mAP and +28\% in R@P90, and the inference speed became 2.4x faster. These combined improvements boosted our user engagements in STL by more than 20\%.
\section{Candidate retrieval (Unified Visual Embedding)} \label{embedding}
Visual embedding is crucial for image retrieval quality. In VPG, the system requires retrieval in two directions: product-to-inspiration, and inspiration-to-product. We assess visual similarity using the Euclidean distance between the embedding representations of the full query image and the candidate objects. In this section, we discuss significant enhancements to our unified visual embedding model since its most recent release \cite{uve3}.
\subsection{Large-scale Pretraining}
Several papers have tried to improve the performance of visual recognition models by scaling up models, including CNNs and Vision Transformers \cite{zhai2022scaling} \cite{mahajan2018exploring}. Scaling up the models, in this context, refers to increasing the size and complexity of the models by adding more layers, increasing the number of parameters or using larger input sizes \cite{xie2017aggregated}. In this work, we describe one of the largest levers for improvement of our extremely similar performance, which comes from scaling the capacity of our transformer-based vision backbone from 250M parameters \cite{uve3} to 1B parameters. The larger model can learn more complex patterns and relationships within the data, leading to a richer and more nuanced understanding of visual attributes such as color, texture, and decorative motifs which help the model to distinguish between exact matches of products versus different products. The key tools that enabled us to scale up the model capacity include Fully Sharded Data Parallel (ZeRO \cite{rajbhandari2020zero}), which optimizes memory usage by distributing model parameters, gradients, and optimizer states across multiple GPUs; Activation Checkpointing, which discards intermediate activations and recomputes them during the backward pass, allowing larger models to fit within available GPU memory; and Flash Attention \cite{dao2022flashattention}, which uses tiling to reduce memory reads and writes between GPU high bandwidth memory (HBM) and on-chip SRAM, resulting in faster attention calculations.
\subsection{Triplets Training Task}
\begin{figure}
    \centering
    \includegraphics[width=0.75\linewidth]{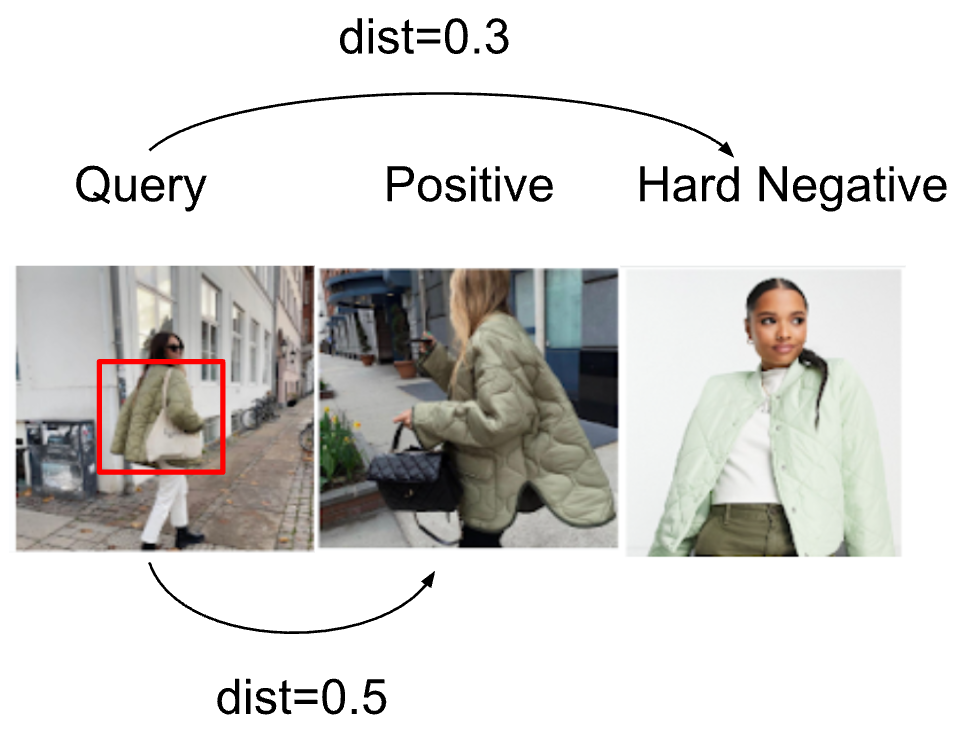}
    \caption{Illustration of a hard triplet scenario: The unified visual embedding model inaccurately estimates the distance between the query and an extremely similar result (positive) as greater than the distance to a less relevant result (negative). A hard triplet is constructed to facilitate the embedding model's learning to rectify this discrepancy.
}
    \label{fig:hard_triplets}
\end{figure}

Triplet training is a concept widely used in metric learning \cite{schultz2003learning} which enforces pairs of images belonging to the same class to have a smaller distance than those of different classes in the embedding space. In our existing STL system, we found several cases where the extremely similar candidate for a given query image was ranked lower than other candidates in the retrieval results, and this problem is well formulated using a triplet loss where we want to reduce the distance between the query image and its extremely similar candidate as compared to other candidates. (Figure \ref{fig:hard_triplets}). In order to improve the model's ability to distinguish between extremely similar candidates and others, we collect a hard triplets dataset, which are triplets where the model initially predicts the negative as being closer to the query than the positive (More details in \ref{hard_triplets_dataset_generation}). Our collected dataset contains approximately 100k triplets, with half being hard triplets and half being random triplets. Home decor and fashion triplets are in this same dataset. 

\subsection{Embedding Representation}
Up until recently \cite{uve3}, we used a binary representation of our embeddings in order to save on storage cost and improve the efficiency of performing the nearest neighbor embedding search. However, with improved mixed precision and low precision training formats, we were able to gain a much improved offline eval performance using 256-dim float vectors versus 1024-dim binary bits. Representing embeddings as floats provides more precision in encoding information from images into the embeddings, and the ability to store floats as half precision floats (16 bits) compared to regular full precision (32 bits) also helps us save on storage costs.

\begin{table}
    \centering
    \scalebox{0.81}{
    \begin{tabular}{l|c|c|c} \hline  
         Model& \begin{tabular}{c} Offline\\VS P@1  \end{tabular}&\begin{tabular}{c}E2E Extremely\\Similar@1 \end{tabular}&\begin{tabular}{c}E2E\\Similar@1 \end{tabular}\\ \hline  
         Production \cite{uve3}& 
     54.7& 23.9&42.3\\ \hline 
 \ ~\ + Large scale pretraining& 64.3& 25.5&67.18\\ \hline  
 \begin{tabular}{l}+ Large scale pretraining\\+ Triplets training\\+ float representations\end{tabular}& $\boldsymbol{70.1}$& $\boldsymbol{27.2}$&$\boldsymbol{74.9}$\\ \hline \end{tabular}}
    \caption{Visual Shopping end-to-end retrieval performance using the latest Unified Embedding model.}
    \label{tab:uve_comparisons}
\end{table}

\begin{table}
    \centering
    \scalebox{0.81}{
    \begin{tabular}{l|c} \hline  
         Model& Product Click-through Rate \\ \hline 
 \ ~\ + Large scale pretraining& +2.57\% \\ \hline  
 \begin{tabular}{l}+ Large scale pretraining\\+ Triplets training\\+ float representations\end{tabular}& +8.59\% \\ \hline \end{tabular}}
    \caption{Visual Shopping end-to-end retrieval performance using the latest Unified Embedding model.}
    \label{tab:uve_ab}
\end{table}

\subsection{Embedding experiments}
With the implementation of the aforementioned enhancements, we have observed substantial improvements in retrieval extremely similar quality, and consequently having a direct impact on user engagements.
\subsubsection{Offline evaluation}
For offline evaluation, we are using the same setup as the one in the \cite{stl} paper. We evaluate the average offline retrieval P@1 from both fashion and home decor to understand how often we get exact product matches given a query. The results are detailed in Table \ref{tab:uve_comparisons}. We observed +28\% extremely similar @ 1 in the offline evaluation with all the improvements.
\subsubsection{Human relevance evaluation}
We show embedding end-to-end relevance results in Table \ref{tab:uve_comparisons} for fashion for popular STL pins in the past 7 days. We observed that the gain is consistent with our offline evaluation findings and we are achieving a +14\% gain for extremely similar @ 1 and +77\% for similar @ 1, which is quite significant. 
\subsubsection{Online A/B experiment}
Because VPG is a new feature, it would be hard to measure the relative impact of the embeddings in this new Ways-to-Style-It module. Therefore, we measure the embedding's impact in the existing STL module and observe the product click-through rate in Table \ref{tab:uve_ab}. We found that user engagements are positively correlated to the significant relevance gains.

\section{Comparison with Complementary Product Recommendations} \label{ctl_comparison}
\begin{figure}
    \centering
    \includegraphics[width=1\linewidth]{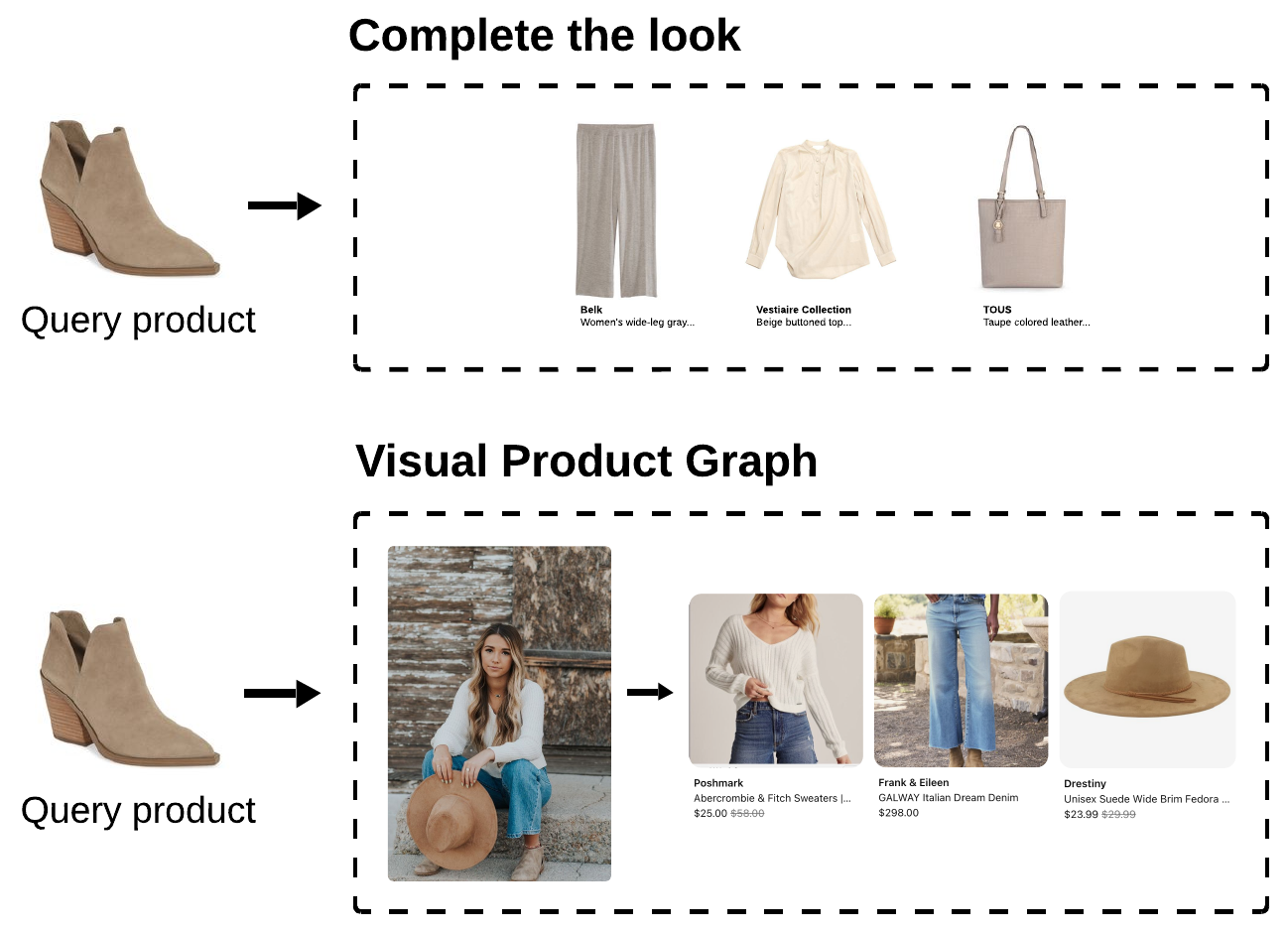}
    \caption{Comparison between our previous "Complete the look" \cite{ctl} approach (top) and our current Visual Product Graph approach (bottom). The former approach's quality was much lower, and it was hard to imagine the recommended products worn together. In contrast, the latter shows all the complementary products together in a single image, and thereby also provides "social proof" that this is a good outfit.}
    \label{fig:comparison}
\end{figure}

The growing practice of complementary product recommendations, often called "Complete the Look," aims to recommend cohesive product sets. Most systems \cite{veit2015learningvisualclothingstyle} \cite{wayfair} suggest items separately, like pants and shoes apart from a shirt. These results can be challenging for users to visualize together, especially if images have plain backgrounds. Our VPG overcomes this by retrieving a single image showcasing the entire outfit, improving the user experience and providing "social proof" by featuring real-world applications of the styles. Initially, we explored a traditional triplet-based recommendation system \cite{ctl} but faced challenges such as scalability, subjective style evaluation, and ineffective visual feature reliance. In contrast, VPG employs advanced techniques for better visual compatibility and integration, and delivers superior recommendations. These recommendations include images that provide social proof, making it an appealing choice for users seeking style inspiration. Since VPG is a new technology, comparing its quality quantitatively with state-of-the-art complementary product recommendations is challenging. Therefore we show qualitative comparisons to our CTL models in Figure \ref{fig:comparison}.

\section{Conclusions}
In this paper, we present the Visual Product Graph, a bidirectional mapping framework designed to connect between individual product images and inspirational context-rich visual content. Our system effectively leverages computer vision technologies to produce insightful image search results and complementary product recommendations rooted in social proof.

Key components of our system include a product-to-scene path and a scene-to-product path, featuring an enhanced object detector and a visual embedding model. The object detector is important to our methodology, allowing for precise scene decomposition and an expanded object index. Recent upgrades have markedly improved our ability to identify shoppable objects, increased object localization accuracy, and minimized false positives, resulting in a more streamlined user experience. Our visual embedding model ensures high-quality retrievals and has been optimized to boost the likelihood of users finding exact matches for their queries.

Our empirical findings demonstrate that enhancements to these models significantly impact user engagement. The system's deployment within Pinterest's "Ways to Style It" module highlights real-world applicability, providing users with seamless access to styled ensembles and decor setups.

This research advances the field of visual search and recommendation systems by offering a scalable solution that builds the bridge between visual products and composite images. In the future, we aim to enhance personalization, ensuring recommendations are closely aligned with users' individual style preferences and past interactions. Additionally, multimodal integration, incorporating textual descriptions and other data forms, will be explored to further enrich search results. Continuous improvements in scalability and efficiency remain top priorities to accommodate a growing user base and data volume.

We hope future practitioners will build upon our work to further explore its application potential and pave the way for innovative solutions that bridge the gap between product discovery and inspiration-driven shopping.

% \begin{acks}
% The authors would like to thank the rest of the ATG-Visual team, especially Chuck Rosenberg for his leadership. We thank Eric Kim for helping support the key-value storage system, and Sergey Malyutin for his leadership in the Warsaw team. We also thank the following cross-functional team members who helped us launch Visual Product Graph to the Ways To Style It module: Sai Xiao, Ai Zhang for backend support;
%   Evan Chen, Cristhian Rendon for API support;
%   Tim Leung, Helen Xu, Annie Won, Cindy Zhang for frontend support;
%   Jiawen Yu for PADS support;
%   Chelsea Pelley for PM support and Erin McGrew Chasteen for TPM support;
%   Steven Eggert, Ophelia Ding for Design support.
% \end{acks}

\bibliographystyle{ACM-Reference-Format}
\bibliography{references}

%%
%% If your work has an appendix, this is the place to put it
\appendix
\section{Appendix}
\subsection{Hard triplets dataset generation}\label{hard_triplets_dataset_generation}
\begin{algorithm}
\caption{Hard triplets dataset generation} 
\label{alg:triplets_dataset_generation}
\begin{algorithmic}
\REQUIRE stl\_log\_data from the last 30 days 
\ENSURE TripletsDataset 
\STATE \textbf{Initialize} $\text{Triplets} \leftarrow \emptyset$ 
\STATE \textbf{Initialize} $\text{TripletsDataset} 
\leftarrow \emptyset$ 
\FORALL{query $q \in \text{stl\_log\_data\_30days}$} 
\STATE Let $C(q)$ be the set of candidates associated with query $q$ 
\IF{$|C(q)| \geq 2$} 
\FORALL{candidates $c\_0, c\_n \in C(q)$} 
\IF{$\text{Engagement}(c\_n) > \text{Engagement}(c\_0)$} 
\STATE Construct triplet $(q, c\_n, c\_0)$ 
\STATE \textbf{Add} $(q, c\_n, c\_0)$ to $\text{Triplets}$ 
\ENDIF 
\ENDFOR 
\ENDIF 
\ENDFOR 
\FORALL{triplet$(q, \text{potential positve}, \text{negative}) \in \text{Triplets}$} 
\STATE Run inference to check if $\text{Distance}(\text{potential positive}) < \text{Distance}(\text{negative})$ 
\IF{True} 
\STATE Human label the triplet as \textit{match} or \textit{no match} 
\ENDIF 
\IF{\textit{match} (potential positive) = \text{“match”} \AND \textit{match} (negative) = \text{“no match”}} 
\STATE \textbf{Add} $(q, \text{potential positive}, \text{negative})$ to $\text{TripletsDataset}$ 
\ENDIF 
\ENDFOR 
\RETURN $\text{TripletsDataset}$
\end{algorithmic}
\end{algorithm}
We outline the construction process for a hard triplets dataset, detailed in Algorithm \ref{alg:triplets_dataset_generation}. The "stl\_log\_data\_30\_days" captures user activity logs in the STL application. In this application, users interact with composite images by clicking on pre-generated dots, which are outputs from our object detection model. Upon clicking a dot, users are shown a "Shop The Look" (STL) feed, recommending visually similar shoppable products.

We log essential details regarding user interactions, recording the query image signature, query box coordinates, query category, candidate image signature, candidate slot, and the count of close-ups on each candidate. This detailed logging allows us to construct informative hard triplets, addressing challenges in visual similarity ranking and enhancing model precision.

\subsection{Object index filtering} \label{object_index_filtering}
\begin{figure}
    \centering
    \includegraphics[width=1\linewidth]{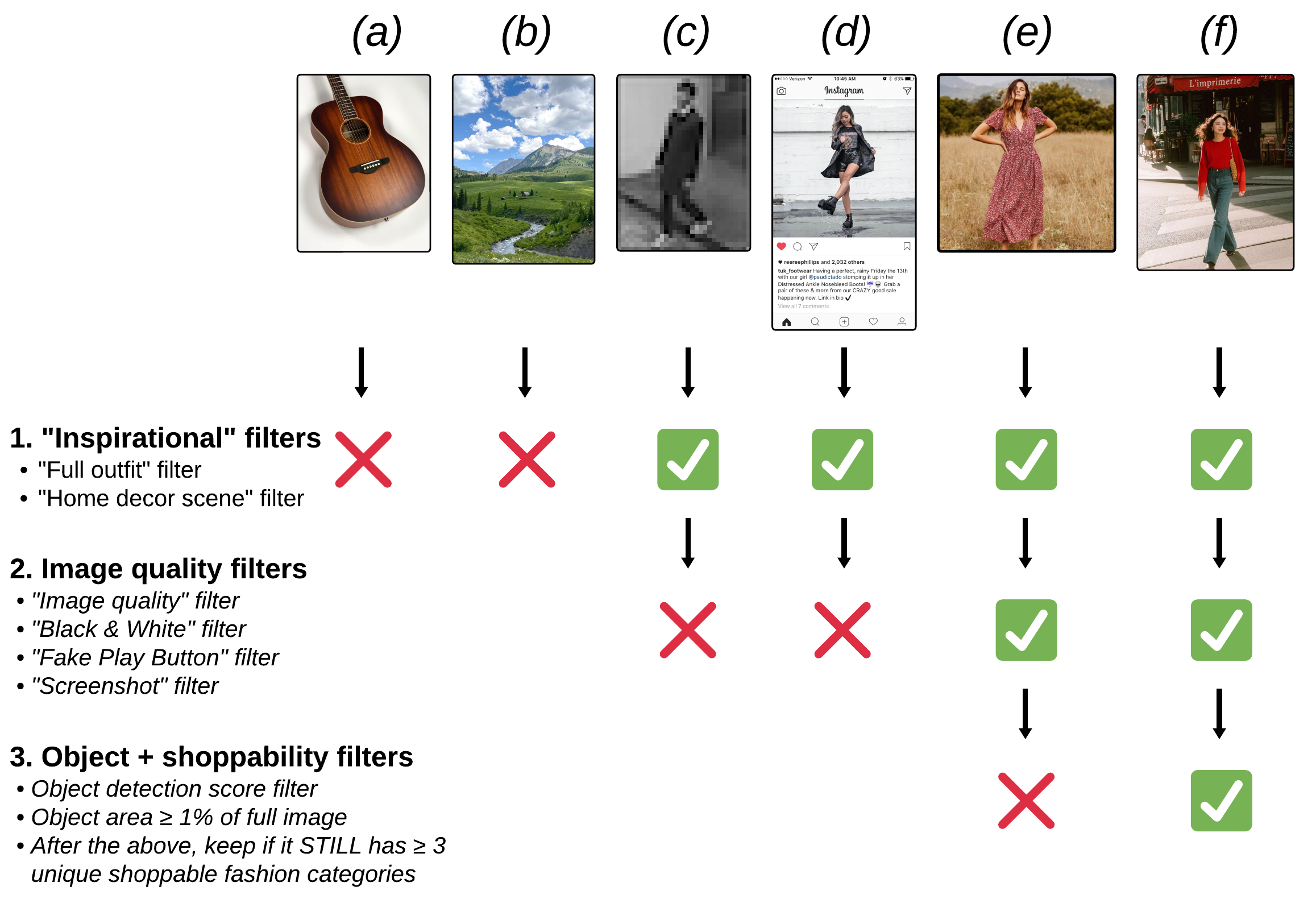}
    \caption{We generate a composite, "inspirational" pin corpus by applying a number of in-house filters to all images on Pinterest.}
    \label{fig:inspo_corpus}
\end{figure}

In the object index for reverse-STL, we ensure that only high-quality inspirational content is kept. The filtering criteria and visualization is shown in Figure \ref{fig:inspo_corpus}.

\begin{figure}
    \centering
    \includegraphics[width=1\linewidth]{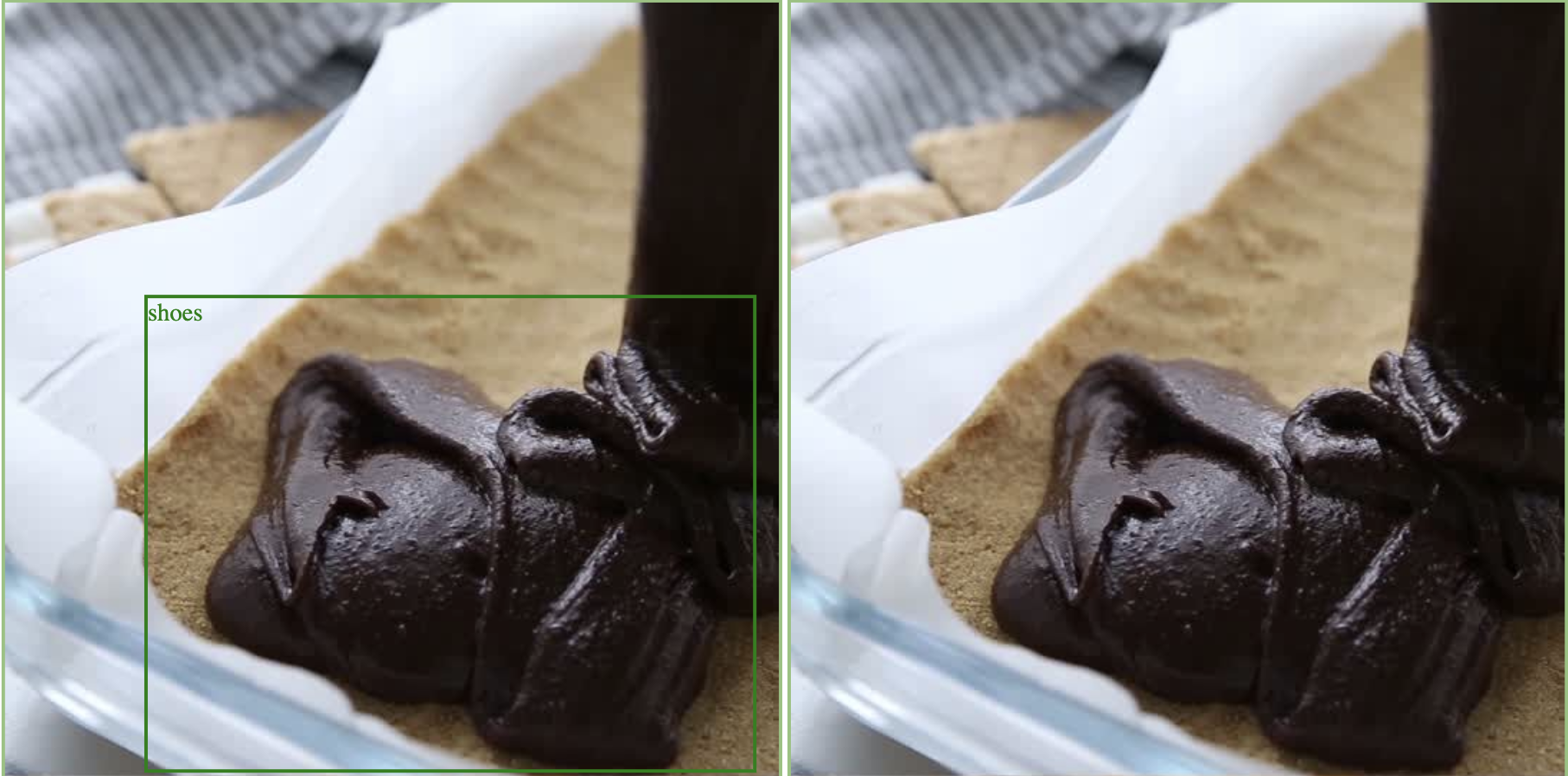}
    \caption{Common failure mode in the object detection step (left: prod): false positives. We address these through a combination of more powerful models and extensive data cleaning.}
    \label{fig:fp}
\end{figure}
\begin{figure}
    \centering
    \includegraphics[width=1\linewidth]{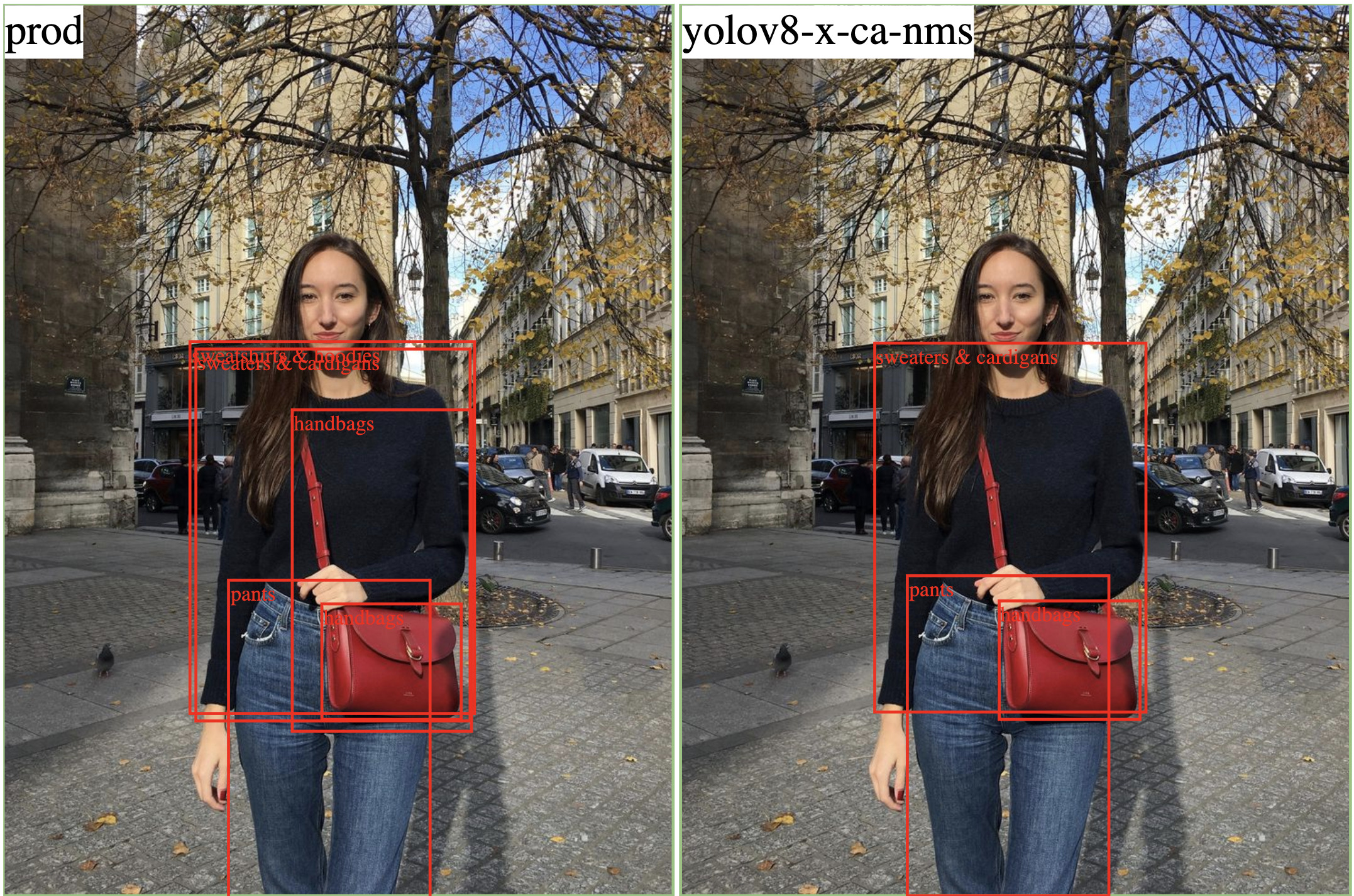}
    \caption{Left: Common duplicate boxes failure mode in the object detection step of a system like Forward-STL / Reverse-STL. Right: Addressed in our new model}
    \label{fig:dup_box}
\end{figure}

\end{document}